\documentclass{article} % For LaTeX2e
\usepackage[dblblindworkshop, final]{neurips_2025}
\workshoptitle{Evaluating the Evolving LLM Lifecycle}

\usepackage{amsthm}

\usepackage{amsmath,amsfonts,bm}

\def\eqref#1{equation~\ref{#1}}
\def\1{\bm{1}}

\DeclareMathAlphabet{\mathsfit}{\encodingdefault}{\sfdefault}{m}{sl}
\SetMathAlphabet{\mathsfit}{bold}{\encodingdefault}{\sfdefault}{bx}{n}

\usepackage{hyperref}
\usepackage{url}

\usepackage[utf8]{inputenc} % allow utf-8 input
\usepackage[T1]{fontenc}    % use 8-bit T1 fonts
\usepackage{hyperref}       % hyperlinks
\usepackage{url}            % simple URL typesetting
\usepackage{booktabs}       % professional-quality tables
\usepackage{amsfonts}       % blackboard math symbols
\usepackage{nicefrac}       % compact symbols for 1/2, etc.
\usepackage{microtype}      % microtypography
\usepackage{xcolor}         % colors

\usepackage{booktabs}

\usepackage{booktabs}
\usepackage{multirow}
\usepackage{makecell}
\usepackage{tabularx}
\usepackage{array}
\usepackage{caption}

\usepackage[utf8]{inputenc}
\usepackage{fancybox}

\usepackage{lipsum}
\usepackage{float}         % for \newfloat
\usepackage{caption}       % caption control (optional but nice)

\usepackage[font=small, skip=0.5ex]{caption}
\usepackage[skip=0.3em, indent]{parskip}

\usepackage{booktabs}
\usepackage{array}
\newcolumntype{L}[1]{>{\raggedright\arraybackslash}p{#1}}
\newcolumntype{C}[1]{>{\centering\arraybackslash}p{#1}}

\usepackage{xspace}

\usepackage{tabularx}
\usepackage{amsthm}
\usepackage{makecell}       % for \makecell and multi-line cells
\usepackage{amsmath}
\usepackage{dsfont}
\usepackage{wrapfig}
\usepackage[table]{colortbl}
\usepackage{makecell}
\usepackage{caption}
\usepackage{multirow}

\usepackage{graphicx}
\usepackage{textcomp}
\usepackage{colortbl}
\usepackage{standalone}
\usepackage{booktabs}
\usepackage[shortlabels]{enumitem}

\definecolor{LightCyan}{rgb}{0.88,1,1}

\usepackage{multirow}
\usepackage{pgfplots}

\usepackage[nolist,nohyperlinks]{acronym}

\usepackage{subcaption}

\usepackage{xspace}
\usepackage{subfiles}
\usepackage{multirow}
\usepackage{tabularx}
\usepackage{subcaption}
\usepackage{enumitem}
\usepackage{balance}
\usepackage{xcolor}
\usepackage{cleveref}
\usepackage[splitrule,bottom]{footmisc}
\usepackage{longtable}
\usepackage{graphicx} 
\usepackage{rotating} 
\usepackage{lipsum}  
\usepackage{float} 
\usepackage{placeins} 
\usepackage{array} 
\usepackage{soul}
\usepackage{makecell}
\usepackage[font=small, skip=0.5ex]{caption}
\usepackage{amsmath}
\usepackage{booktabs} 
\usepackage{pifont}
\usepackage{longtable}
\usepackage{multirow}
\usepackage{array}

\title{A Systematic Evaluation of Preference Aggregation in Federated RLHF for Pluralistic Alignment of LLMs}

\author{Mahmoud Srewa, Tianyu Zhao, \& Salma Elmalaki \\
Department of Electrical Engineering and Computer Science\\
University of California, Irvine\\
Irvine, CA 92697, USA \\
\texttt{\{msrewa, tzhao15, salma.elmalaki\}@uci.edu} 
}

\begin{document}

\begin{acronym}
    \acro{har}[HAR]{Human Activity Recognition}
    \acro{llms}[LLMs]{Large Language Models}
    \acro{llm}[LLM]{Large Language Model}
    \acro{fl}[FL]{Federated Learning}
    \acro{gpo}[GPO]{Group Preference Optimization}
    \acro{rlhf}[RLHF]{Reinforcement Learning from Human Feedback}
    \acro{dpo}[DPO]{Direct Preference Optimization}
    \acro{ppo}[PPO]{Proximal Policy Optimization}
\end{acronym}

\maketitle

\begin{abstract}
This paper addresses the challenge of aligning \ac{llms} with diverse human preferences within \ac{fl} environments where standard methods often fail to adequately represent diverse viewpoints.  
We introduce a comprehensive evaluation framework that systematically assesses the trade-off between alignment quality and fairness when using different aggregation strategies for human preferences. In our federated setting, each group locally evaluates rollouts and produces reward signals, and the server aggregates these group-level rewards without accessing any raw data.
Specifically, we evaluate standard reward aggregation techniques (min, max, and average) and introduce a novel adaptive scheme that dynamically adjusts preference weights based on a group's historical alignment performance. Our experiments on Q/A tasks using a \ac{ppo}-based \ac{rlhf} pipeline demonstrate that our adaptive approach consistently achieves superior fairness while maintaining competitive alignment scores. This work offers a robust methodology for evaluating \ac{llm} behavior across diverse populations and provides a practical solution for developing truly pluralistic and fairly aligned models.
\end{abstract}

\section{Introduction}\label{intro}

The remarkable capabilities of \ac{llms} have positioned them as a central technology across various domains. However, their real-world utility and safety hinge on their ability to align with complex and diverse human values and social norms\cite{intro1, intro2}. The prevailing methodology for this alignment is \ac{rlhf}, which fine-tunes models based on collected human preference data~\cite{intr3-rlhf}. While effective, the standard \ac{rlhf} paradigm often operates on a centralized dataset, which is not only a privacy concern but also risks embedding biases of a narrow demographic~\cite{intro4}.

To address this, the integration of \ac{rlhf} with \ac{fl} has emerged as a promising avenue. \ac{fl} allows for model training on decentralized data from numerous clients, thus preserving data privacy and capturing a wider range of human preferences \cite{llmfed,srewa2025pluralllm}. However, this fusion presents a critical and underexplored challenge: \textbf{How to aggregate the diverse and potentially conflicting preference signals from different user groups?} In our setting, these preference signals appear as per-group reward scores generated locally by each client; the server aggregates these decentralized reward vectors without accessing any raw data to compute a global reward used for PPO updates.
% \srewa{In our setting, these preference signals appear as per-group reward scores generated locally by each client; the server aggregates these decentralized reward vectors—without accessing any raw data—to compute a global reward used for PPO updates.}
The choice of aggregation strategy is not merely a technical detail; it is an evaluation protocol that directly shapes the model's final behavior, determining whose preferences are prioritized and whose are marginalized.

This paper proposes a systematic evaluation framework to analyze the impact of different aggregation techniques on both alignment performance and fairness. By comparing standard reward aggregation methods with our proposed adaptive aggregation scheme, our goal is to define a more robust protocol to assess LLMs in decentralized, pluralistic environments. We show that while simple aggregation methods can lead to unintended biases, our adaptive approach strikes a superior balance between achieving strong overall alignment and ensuring equitable representation across diverse groups, thus contributing to the development of more reliable and justly aligned LLMs. Our approach follows a zero-shot alignment paradigm, using only aggregated group reward signals without task demonstrations, ensuring generalizable alignment.
\section{Background and Related Work}\label{sec:related}

% The alignment of \ac{llms} with complex human preferences is a central goal in their development. Since explicitly defining human values in a loss function is challenging, a robust paradigm has emerged in which models learn directly from human preference data. \ac{rlhf} has become a powerful technique for this purpose, guiding LLMs toward desired behaviors like safety and helpfulness~\cite{intr3-rlhf}. The most common RL algorithm used in \ac{rlhf} is \ac{ppo} which fine-tunes the LLM by receiving feedback from a reward model trained on human preference pairs~\cite{schulman2017ppo}. An alternative, Direct Preference Optimization (DPO), simplifies this process by bypassing the need for a separate reward model, instead directly optimizing the model to assign a higher probability to preferred responses~\cite{rafailov2023direct}.

The alignment of \ac{llms} with complex human preferences is a central goal in their development. Because explicitly encoding human values into a fixed loss function is challenging, \emph{Reinforcement Learning from Human Feedback} (\ac{rlhf}) has emerged as the dominant paradigm for steering models toward desired behavior. In \ac{rlhf}, models learn from human preference data via a reward model trained on pairwise comparisons, and the policy is then optimized to maximize this learned reward~\cite{intr3-rlhf}. The most commonly used RL algorithm in \ac{rlhf} is \ac{ppo}, which fine-tunes the LLM using feedback from the reward model trained on human preference pairs~\cite{schulman2017ppo}. An alternative, Direct Preference Optimization (DPO), simplifies this pipeline by bypassing an explicit reward model and directly optimizing the policy to assign higher probability to preferred responses~\cite{rafailov2023direct}.

While traditional alignment methods typically aim for a single, global preference objective, real-world users exhibit diverse and sometimes conflicting preferences. Group Preference Optimization (GPO)~\cite{gpo} was introduced to address this challenge by enabling group-specific alignment of LLMs. GPO augments the base LLM with a lightweight transformer module trained via in-context supervised learning to predict and incorporate the distinct preferences of user groups using only a few examples. This auxiliary module serves as a preference predictor that captures alignment patterns across heterogeneous communities, allowing the model to dynamically adapt its responses according to different social or demographic contexts. In parallel, methods such as Group Robust Policy Optimization (GRPO)~\cite{ramesh2024group} and MaxMin-RLHF~\cite{maxminrlf} have focused on ensuring robustness across diverse user populations within centralized \ac{rlhf} settings. However, these methods still rely on collecting and processing user data on a central server, which raises privacy and data ownership concerns. To overcome these limitations,

To overcome these limitations, PluralLLM~\cite{srewa2025pluralllm} extends GPO \cite{gpo} into a federated learning architecture that enables groups to collaboratively learn lightweight transformer-based preference predictors without sharing raw data. Concretely, each group trains a lightweight transformer using few-shot in-context examples in a \ac{fl} manner, producing a local preference module that can predict, for any Q/A question, a probability distribution over all answer options reflecting that group's latent preferences. This module serves as a fully local reward model; given \ac{rlhf} rollouts from the server, the PluralLLM outputs group-specific preference probabilities that can be transformed into scalar rewards.

Our work builds directly on this foundation by using these PluralLLM predictors as decentralized reward generators for each group. At each \ac{ppo} iteration, the server broadcasts rollouts to all groups, each group evaluates the responses using its PluralLLM module, and returns group-specific reward vectors. The central question we address is how to aggregate these heterogeneous and sometimes conflicting reward signals across groups. By systematically comparing standard aggregation schemes and introducing a dynamic alpha aggregation strategy, we analyze how different reward-aggregation protocols affect both fairness and alignment quality in multi-group federated \ac{rlhf}.

\section{Methodology}\label{sec:method}

\begin{figure*}[!t]
\centering
\includegraphics[width=1\textwidth]{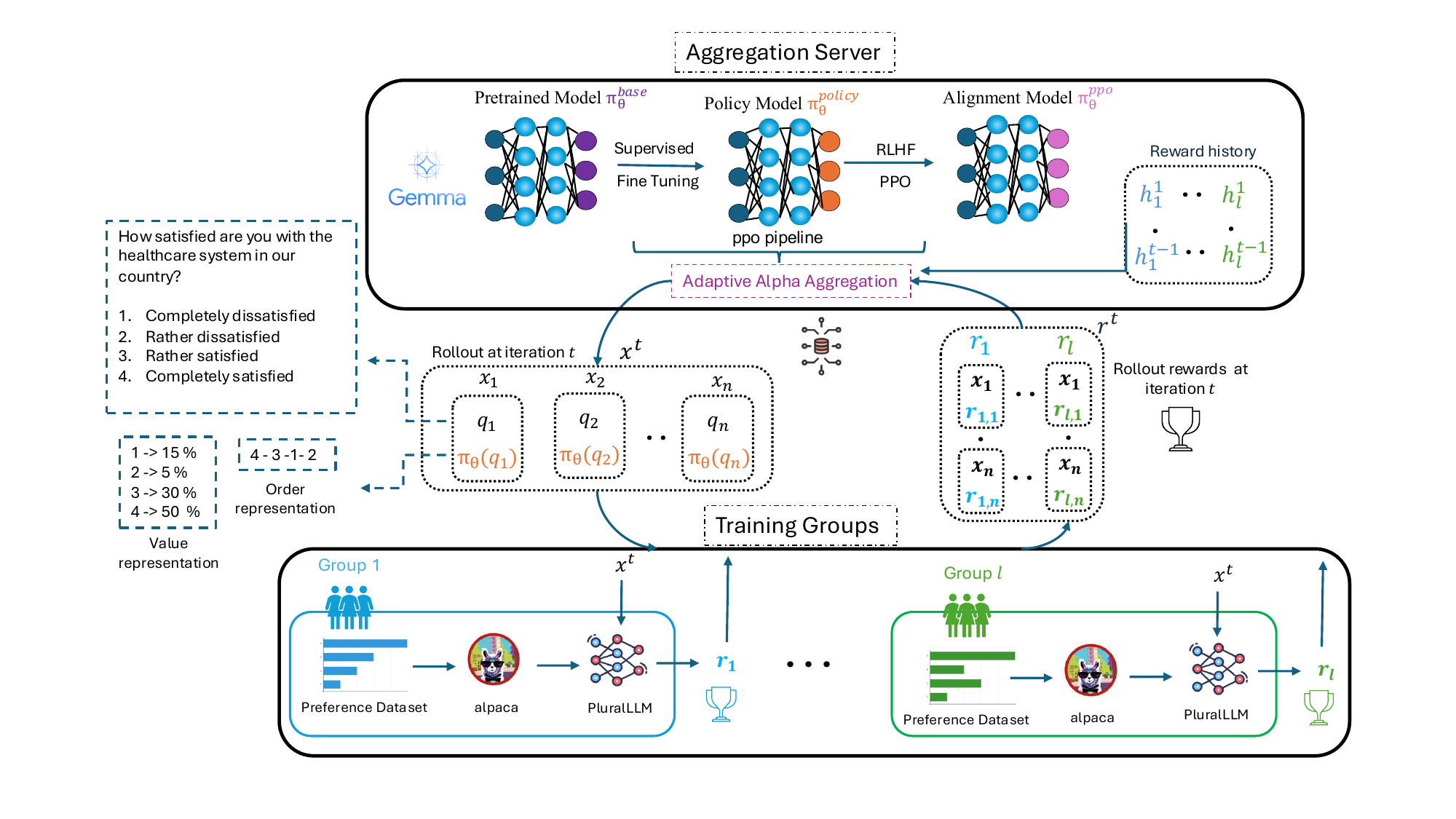}
 \caption{Federated \ac{rlhf} for pluralistic alignment of group preferences in LLM.  }
\label{fig:framework}
\end{figure*}

\paragraph{System Setup and Training Groups:}

% \srewa{In our setting, each group $g_i$ corresponds to a distinct demographic or preference cluster (e.g., age, region, political leaning), and each group acts as a single federated client that locally represents its users' aggregated preferences.}
In our setting, each group $g_i$ corresponds to a distinct demographic or preference cluster (e.g., age, region, political leaning), and each group acts as a single federated client that locally represents its users' aggregated preferences.
Our framework focuses on Q/A tasks with  $l$ training groups $G_{\text{train}} = \{g_1, g_2, \ldots, g_l\}$, where each group $g_i$ maintains its private preference dataset $D_{g_i} = \{(x_{i,j}, y_{i,j})\}$ locally. Each preference sample consists of a query-response pair embedding $x_{i,j}$ and the corresponding group preference probability $y_{i,j}$. These datasets are distributed across groups and never shared with the central server, ensuring privacy preservation.

%\subsection{System Flow and Model Training}

As illustrated in~\autoref{fig:framework}, the aggregation server is initialized with a base LLM model $\pi^{base}_{\theta}$ and performs supervised fine-tuning (SFT) to adapt it for Q/A tasks, resulting in a policy model $\pi^{policy}_{\theta}$ suitable for PPO training. %The system operates in an iterative manner where 
The server coordinates between policy optimization and distributed preference learning. 

At iteration $t$, the server generates rollouts $X^t$ consisting of queries (questions with multiple choice's options) and LLM responses using the current policy $\pi^{policy}_{\theta}$. These rollouts are distributed to all training groups for preference evaluation. 
% Each group $g_i$ uses the PluraLLM GPO model \cite{srewa2025pluralllm} alongside their local preference dataset $D_{g_i}$ as context points to generate preference probabilities for the received rollouts.

\paragraph{Distributed Reward Generation:}

Each group $g_i$ first uses the PluraLLM \cite{srewa2025pluralllm} as a local lightweight reward model to generate preference probabilities for the received rollouts at iteration $t$. %Based on the specific task, 
These probabilities are then converted to rewards $r^t_{g_i}$ according to the reward metric used. In our evaluation, we focus on two approaches: (1) preference probability prediction, where rewards are calculated directly from the predicted probabilities, and (2) preference ranking, where the probabilities are first converted to rankings before reward calculation. These task-specific rewards $r^t = \{r^t_{g_1}, r^t_{g_2}, \ldots, r^t_{g_l}\}$ reflecting how well the generated responses align with each group's preferences, are transmitted back to the aggregation server.

% \paragraph{Adaptive Alpha Aggregation:} The core of the federated \ac{rlhf} framework is the aggregation of local models. At each training round, the central server receives rewards $r^t$ from different groups and combines them to form a new global reward model. We evaluated three standard aggregation methods and proposed a novel adaptive scheme.
\paragraph{Adaptive Alpha Aggregation:}
The core of the federated \ac{rlhf} framework is the aggregation of local rewards. At each training round, the central server receives per-group rewards $r^t$ from different groups and aggregates them into a final global reward $Agg_\alpha(\mathbf{r}^t)$, ready to be used for updating the policy. We have $r^t=\{ r^t_{g_1}, r^t_{g_2},r^t_{g_3}, \dots, r^t_{g_l}\}$ from $l$ clients, where $r^t_i$ represents how well the LLM outputs align with group  $g_i$ preference at \ac{fl} iteration $t$. Recent work in the literature introduced an aggregation method, namely alpha aggregation\cite{park2024rlhf} which achieves the consensus among heterogeneous feedback in RLHF. This is highlighted in \autoref{eq: consensus aggregation}. This consensus reward aggregation is controlled by $\alpha$, $Agg_\alpha(\textbf{r}) = max(\textbf{r})$, when $\alpha = \infty$ and $Agg_\alpha(\textbf{r}) = min(\textbf{r})$, when $\alpha = -\infty$.

\begin{equation}\label{eq: consensus aggregation}
    Agg_\alpha(\textbf{r}) = 
    \begin{cases}
    \frac{1}{\alpha}\log(\frac{1}{N}\sum_{i \in N}\exp(\alpha r_i)) &\alpha \neq 0 \\
    \frac{1}{N}\sum_{i \in N}r_i &\alpha= 0
    \end{cases}
\end{equation}

% We propose an adaptive aggregation technique that uses this alpha aggregation. %In particular, 
We propose an \emph{adaptive} extension of this scheme that replaces the single global $\alpha$ with \emph{group-specific}, dynamically updated weights $\alpha_g^t$. Instead of using one fixed $\alpha$ for all clients, our method learns $\alpha_g^t$ per group based on its historical alignment performance, and plugs these into the alpha aggregation:

\begin{equation}\label{eq: adaptive aggregation}
    Agg_\alpha(\textbf{r}^t) = \begin{cases}
\frac{1}{|G_{\text{train}}|} \sum_{g \in G_{\text{train}}} r^t_{g} & \text{if } FI \ge 0.9 \\
\log\left(\frac{1}{|G_{\text{train}}|} \sum_{g \in G_{\text{train}}} \exp(\alpha_g^t \cdot r^t_{g})\right) & \text{otherwise}
\end{cases} 
\end{equation} 
To achieve a balanced accumulated alignment reward history $\textbf{h} = \{h_{g_1}, h_{g_2}, h_{g_3}, \dots, h_{g_l}\}$, across clients, the aggregation weights $\alpha_i$ are dynamically adjusted in inverse proportion to each client's historical alignment performance($\alpha_i = \texttt{softmax}(1-h_i)$). Specifically, a client $i$ with a lower accumulated alignment reward, $h_i$, is assigned a higher weight, $\alpha_i$. As shown in \autoref{eq: adaptive aggregation}, a higher $\alpha_i$ value increases the dominance of the corresponding reward $r_i$. Hence, the $\alpha_i$ value for client $i$ changes adaptively based on the alignment history for this client $h_i$. 

\paragraph{Fairness Index ($FI$):}
To determine when uniform averaging is sufficient versus when adaptive weighting is needed, 
we use a $FI$ that measures the similarity of per-group rewards 
$r_{g_i}^t$ across all groups. $FI$ ranges from 0 to 1, where values near 1 indicate 
highly consistent (fair) rewards and lower values indicate increasing disparity. 
When $FI \approx 1$ (we use a practical threshold $FI \ge 0.9$), the rewards across 
groups are nearly uniform, and thus a simple average aggregation is appropriate. 
Otherwise, we employ the $\alpha$-weighted log-sum-exp aggregation to amplify 
contributions from groups with historically lower alignment performance.
% ---- END INSERTED PARAGRAPH ----

% Upon receiving rewards from all groups, the server executes our adaptive alpha aggregation algorithm that considers both historical alignment scores $AS_t^{g_i}$ and fairness metrics across groups. The algorithm dynamically adjusts weights $\alpha_{g_i}^t$ to favor groups with lower historical performance, ensuring equitable representation without leaving any group behind. The aggregated reward is computed as:

% \begin{equation}
% r_{\text{final}}^t = \sum_{i=1}^{l} \alpha_{g_i}^t \cdot r_{g_i}^t
% \end{equation}

% where $\alpha_{g_i}^t$ are the adaptive weights that balance between alignment quality and fairness (see Appendix for algorithm formalization).

\paragraph{PPO Training and Iteration:}

With the aggregated rewards $Agg_\alpha(r^t)$, the server performs PPO optimization to update the policy model $\pi^{policy}_{\theta^t} \rightarrow \pi^{policy}_{\theta^{t+1}}$. The updated policy generates new rollouts for the next iteration, and the process continues until a predefined number of iterations or specific alignment score is reached. This iterative approach ensures continuous adaptation to diverse group preferences while maintaining fairness through our adaptive aggregation scheme.
\section{Evaluation}\label{sec:eval}

We evaluate our approach using Gemma-2B-it, a fine-tuned version of the Gemma model, as our base LLM \cite{gemma}, More details on the experiment setup and configuration are summarized in Appendix~\ref{appendix:setup}. 

Our experiments utilize the Pew Research Center's Global Attitudes Surveys dataset~\cite{globalopnion}, which captures diverse public opinions across social, political, and economic issues from various demographic groups. The dataset consists of multiple-choice survey questions answered by participants from a wide range of countries, with 2{,}554 questions spanning topics such as politics, media, technology, religion, race, and ethnicity. For each question and country, the dataset provides a probability vector over the answer choices, indicating the fraction of respondents in that country selecting each option. These probability vectors differ across countries, reflecting distinct group-level preferences. In our experiments, we treat each country as a separate user group (i.e., a federated client) and use all available groups in the survey. Our goal is to align the LLM with these diverse group preferences in a fair and robust manner, without overfitting to any single group or majority and without leaving any group behind.

We assess performance using fairness index $FI$ and alignment scores across two primary tasks: 
the \textit{preference probability prediction task} (see Figure~\ref{fig:prob-prompt}) and 
the \textit{preference ranking task} (see Figure~\ref{fig:rank-prompt}). 
Our evaluation framework encompasses various reward functions and aggregation strategies. 
We compare our adaptive alpha aggregation against standard federated approaches (Min, Max, Average) 
and a supervised fine-tuning (SFT) baseline.

\begin{figure}[h] 
\centering
\shadowbox{%
\begin{minipage}{0.9\textwidth}
\textbf{Preference Probability Prediction Prompt}

\hrule
\vspace{5pt}

\texttt{<bos><start\_of\_turn>user}

You are an expert in modelling group preferences.  
You will receive \textbf{a question and exactly 4 options}.

\textbf{Your task}
\begin{itemize}
  \item Assign a preference score to \textbf{each and every option}
  \item Produce \textbf{4} scores—no option may be skipped or combined
  \item Each score must be a decimal between 0 and 1, and \textbf{the rounded scores must sum to 1.00}
  \item Higher scores represent options a typical group is more likely to choose
\end{itemize}

\textbf{Output format}
\begin{itemize}
  \item One line, comma-separated decimal numbers, \textbf{no spaces}
  \item Round each to \textbf{2} decimal places
  \item \textbf{No extra text, labels, or symbols}
  \item Example: 0.65,0.20,0.10,0.05
\end{itemize}

Return ONLY the 4 scores in the same order as options.

\textbf{Question:} Germany’s influence in the EU  
Options:  
A: Has too much influence  
B: Has too little influence  
C: Has about the right amount of influence  
D: DK/Refused

\texttt{<end\_of\_turn>}  
\texttt{<start\_of\_turn>model}
\end{minipage}%
}
\caption{Preference Probability Prediction Prompt}
\label{fig:prob-prompt}
\end{figure}

% --- Option Preference Ranking Prompt ---
\begin{figure}[!h]
\centering
\shadowbox{%
\begin{minipage}{0.9\textwidth}
\textbf{Preference Ranking Prompt}

\hrule
\vspace{5pt}

\texttt{<bos><start\_of\_turn>user}

You are an expert in ranking group preferences.  
You will receive \textbf{a question and exactly 4 options}.

\textbf{Your task}
\begin{itemize}
  \item Rank all \textbf{4 provided options} from most to least preferred
  \item Process every option—no skipping or combining
  \item Order options based on what a typical group would most likely choose
  \item Higher preference options appear first
\end{itemize}

\textbf{Output format}
\begin{itemize}
  \item One line, comma-separated option letters, \textbf{no spaces}
  \item Use the exact provided letters
  \item \textbf{No extra text, labels, or symbols}
  \item Example: B,C,A,D
\end{itemize}

Return ONLY the 4-letter ranking.

\textbf{Question:} Germany’s influence in the EU  
Options:  
A: Has too much influence  
B: Has too little influence  
C: Has about the right amount of influence  
D: DK/Refused

\texttt{<end\_of\_turn>}  
\texttt{<start\_of\_turn>model}
\end{minipage}%
}
\caption{Preference Ranking Prompt}
\label{fig:rank-prompt}
\end{figure}

\subsection{Evaluation Framework}

% The LLM rollout at iteration $t$, denoted $X^t$, consists of questions $\{q_j\}$ and corresponding responses generated by the policy model $\pi_\theta(q_j)$ for each question $q_j$. We parse each response to extract the relevant output and denote the parsed LLM answer as $y^{t,\text{llm}}_j$ for question $j$ at \ac{fl} iteration $t$. Let $o_{j,k}$ be the $k$-th option for question $j$, and let $K$ denote the total number of answer options for that question. For each group $g \in G_{\text{train}}$, we compute rewards $r_{g,j}$ by comparing the LLM output $y^{t,\text{llm}}_j$ against the group $g$ PluralLLM-derived target distribution $p^{\text{PluralLLM}}_{g,j}$~\cite{srewa2025pluralllm}, i.e., the group-specific probability vector over options for question $j$ predicted by the local PluralLLM reward model.

The LLM rollout at \ac{fl} iteration $t$, denoted $X^t$, consists of a set of
questions $\{q_j\}$ together with corresponding responses generated by the
policy model $\pi_\theta$ for each question $q_j$. We parse each response to
extract the relevant output and denote the resulting LLM prediction as
$y^{t,\text{llm}}_j$ for question $j$ at iteration $t$. Let $o_{j,k}$ be the
$k$-th answer option for question $j$, and let $K$ denote the total number of
answer options for that question. For each group $g \in G_{\text{train}}$, we
compute a reward $r_{g,j}$ by comparing the LLM prediction $y^{t,\text{llm}}_j$
against the PluralLLM-derived target distribution for that group,
$p^{\text{PluralLLM}}_{g,j}$~\cite{srewa2025pluralllm}, i.e., the
group-specific probability vector over options for question $j$ predicted by
the local PluralLLM reward model.

\subsection{Reward Metrics}

Our evaluation employs two categories of reward metrics to assess alignment
quality across different aspects of preference modeling. These rewards are
also used as alignment scores in our evaluation, i.e., they directly define
the Avg AS and Min AS reported in \autoref{tab:fairness-alignment}.

\subsubsection{Distance-Based Reward Metrics (Preference Prediction Task)}

These rewards quantify the alignment between the LLM-predicted and PluralLLM-target probability distributions. 
They are computed locally on each client in $G_{\text{train}}$, where each group $g \in G_{\text{train}}$ 
independently evaluates the following reward functions:

\textbf{Wasserstein Reward:} Measures optimal transport cost between distributions.
\begin{equation}
r_{g,j}^{t,\text{Was}} = \frac{W_1(y^{t,\text{llm}}_j, p^{\text{PluralLLM}}_{g,j})}{K-1}
\end{equation}
$r_{g,j}^{t,\text{Was}} \in [0, 1]$, where 0 indicates a perfect distribution match. Lower values indicate better alignment between LLM and group preferences.

\textbf{Cosine Similarity Reward:} Captures directional similarity between preference vectors.
\begin{equation}
r_{g,j}^{t,\text{Cos}} = \frac{y^{t,\text{llm}}_j \cdot p^{\text{PluralLLM}}_{g,j}}{||y^{t,\text{llm}}_j|| \cdot ||p^{\text{PluralLLM}}_{g,j}||}
\end{equation}
$r_{g,j}^{\text{Cos}} \in [-1, 1]$, where 1 indicates identical direction, 0 indicates orthogonal, and -1 indicates opposite direction. Higher values indicate better preference alignment.

\textbf{KL Divergence Reward:} Measures information-theoretic alignment.
\begin{equation}
r^{t,\mathrm{KL}}_{g,j}
    = D_{\mathrm{KL}}\!\left(p^{\text{PluralLLM}}_{g,j} \,\big\|\, y^{t,\text{llm}}_{j}\right)
    = \sum_{k} p^{\text{PluralLLM}}_{g,j,k} \,
      \log \frac{p^{\text{PluralLLM}}_{g,j,k}}{y^{t,\text{llm}}_{j,k}} .
\end{equation}

$r_{g,j}^{\text{KL}} \in [0,\infty)$, where 0 indicates identical distributions, and more positive = greater divergence. Smaller values indicate better alignment.

\subsubsection{Ranking-Based Reward Metrics (Preference Ranking Task)}

These rewards evaluate preference ordering consistency:

\textbf{Kendall Tau Reward:} Measures rank correlation between LLM and PluralLLM orderings.
% \begin{equation}
% r_{g_i}^{\text{Ken}} = \tau(\text{rank}(y^{\text{llm}}_i), \text{rank}(p^{\text{PluralLLM}}_i))
% \end{equation}
\begin{equation}
r^{t,\text{Ken}}_{g,j}
    = \tau\!\left(
        \text{rank}\!\left(y^{t,\text{llm}}_{j}\right),
        \text{rank}\!\left(p^{\text{PluralLLM}}_{g,j}\right)
      \right) .
\end{equation}

$r_{g,j}^{t,\text{Ken}} \in [-1, 1]$, where 1 indicates perfect rank agreement, 0 indicates no correlation, and -1 indicates perfect disagreement. Higher values indicate better ranking alignment.

% \textbf{Borda Reward:} Position-weighted scoring based on ranking accuracy.
% % \begin{equation}
% % r_{g_i}^{\text{Bor}} = \frac{\sum_{j=1}^n (n-j+1) \cdot \mathbb{I}[\text{rank}(y^{\text{llm}}_i)_j = \text{rank}(p^{\text{PluralLLM}}_i)_j]}{n(n+1)/2}
% % \end{equation}
% \begin{equation}
% r^{t,\text{Bor}}_{g,j}
%     = \frac{\displaystyle\sum_{k=1}^{n} (n - k + 1)\,
%         \mathbb{I}\!\Big[
%             \text{rank}\!\big(y^{t,\text{llm}}_{j}\big)_k
%             =
%             \text{rank}\!\big(p^{\text{PluralLLM}}_{g,j}\big)_k
%         \Big]
%       }{\displaystyle n(n+1)/2}
% \end{equation}

% $r_{g,j}^{t,\text{Bor}} \in [0, 1]$, where 1 indicates perfect position-wise ranking match, and 0 indicates no correct positions. Higher values indicate better ranking quality.

\textbf{Borda Reward:} Position-weighted scoring based on ranking accuracy.
\begin{equation}
r^{t,\text{Bor}}_{g,j}
    = \frac{\displaystyle\sum_{k=1}^{K} (K - k + 1)\,
        \mathbb{I}\!\Big[
            \text{rank}\!\big(y^{t,\text{llm}}_{j}\big)_k
            =
            \text{rank}\!\big(p^{\text{PluralLLM}}_{g,j}\big)_k
        \Big]
      }{\displaystyle K(K+1)/2}
\end{equation}
$r_{g,j}^{t,\text{Bor}} \in [0, 1]$, where 1 indicates perfect position-wise ranking match, and 0 indicates no correct positions. Higher values indicate better ranking quality.

\textbf{Binary Reward:} Simple correctness indicator.
\begin{equation}
r_{g,j}^{t,\text{Bin}} = \mathbb{I}[\text{rank}(y^{t,\text{llm}}_j) = \text{rank}(p^{\text{PluralLLM}}_{g,j})]
\end{equation}
$r_{g,j}^{t,\text{Bin}} \in \{0, 1\}$, where 1 indicates exact ranking match, 0 indicates any disagreement. Binary indicator of perfect alignment.

\subsection{Aggregation Schemes}

% Let $G_{\text{train}} = \{g_1, g_2, \ldots, g_l\}$ be the set of training groups. For each question $q_i$, the server aggregates the rewards $\{r_{g_1}, r_{g_2}, \ldots, r_{g_l}\}$ across groups using different strategies:
For each question $q_j$ at \ac{fl} iteration $t$, the server aggregates the client-level rewards
$\{r^t_{g_1,j}, r^t_{g_2,j}, \ldots, r^t_{g_l,j}\}$ across groups using different strategies:

\textbf{Average Aggregation:}
\begin{equation}
r_{\text{final},j}^t = \frac{1}{|G_{\text{train}}|} \sum_{g \in G_{\text{train}}} r^t_{g,j}
\end{equation}
Provides balanced representation but may mask group-specific needs.

\textbf{Min Aggregation:}
\begin{equation}
r_{\text{final},j}^t = \min_{g \in G_{\text{train}}} r^t_{g,j}
\end{equation}
Ensures no group is left behind but may be overly conservative, limiting overall performance.

\textbf{Max Aggregation:}
\begin{equation}
r_{\text{final},j}^t = \max_{g \in G_{\text{train}}} r^t_{g,j}
\end{equation}
Optimizes for best-case performance but may neglect underrepresented groups.

\textbf{Adaptive Alpha Aggregation:}
\begin{equation}
r_{\text{final},j}^t = \begin{cases}
\frac{1}{|G_{\text{train}}|} \sum_{g \in G_{\text{train}}} r^t_{g,j} & \text{if } FI \ge 0.9 \\
\log\left(\frac{1}{|G_{\text{train}}|} \sum_{g \in G_{\text{train}}} \exp(\alpha_g^t \cdot r^t_{g,j})\right) & \text{otherwise}
\end{cases}
\end{equation}
Dynamically balances fairness and performance by favoring historically underperforming groups.

The adaptive weights $\alpha_g^t$ are computed using reversed softmax on historical alignment scores:
\begin{equation}
\alpha_g^t = \frac{\exp((1 - h^{t-1}_g)/T)}{\sum_{g' \in G_{\text{train}}} \exp((1 - h^{t-1}_{g'})/T)}
\end{equation}
with temperature $T = 0.1$ and $h^{t-1}_g$ being group $g$'s historical alignment score.

% We provide an analysis and theoretical proof for convergence with the adaptive alpha aggregation in Appendix~\ref{app:proof}. 

\subsection{Fairness Evaluation Metrics}

The Fairness Index (FI) measures reward variation across groups for the same question-response pair:

% \begin{equation}
% FI = \frac{1}{|X|} \sum_{q \in X
% } \frac{1}{1 + \text{CoV}^2(q)}
% \end{equation}

% where the coefficient of variation for the question-response pair $q_i$ is:
% \begin{equation}
% \text{CoV}(q_j) = \frac{\sigma(\{r^t_{g,j}\}_{g \in G_{\text{train}}})}{\mu(\{r^t_{g,j}\}_{g \in G_{\text{train}}})}
% \end{equation}

\begin{equation}
FI = \frac{1}{|X|} \sum_{q_j \in X} \frac{1}{1 + \text{CoV}^2(q_j)}
\end{equation}

where the coefficient of variation for question $q_j$ is:
\begin{equation}
\text{CoV}(q_j) = 
    \frac{\sigma\big(\{r^t_{g,j}\}_{g \in G_{\text{train}}}\big)}
         {\mu\big(\{r^t_{g,j}\}_{g \in G_{\text{train}}}\big)}.
\end{equation}

$FI \in [0, 1]$, where 1 = perfect fairness (identical rewards across groups), and 0 = maximum unfairness. Higher $FI$ values indicate more equitable treatment across demographic groups, while lower values suggest systematic bias favoring certain groups over others. We compute $FI$ separately for each reward metric (Wasserstein, Cosine, KL, Kendall, Borda, Binary) to characterize fairness under different alignment criteria.

%%%%%%% All results 
\newcolumntype{Y}{>{\raggedleft\arraybackslash}X}
\newcolumntype{L}{>{\raggedright\arraybackslash}X}
\newcommand{\na}{\textemdash} % for N/A cells

\begin{table*}[t]
\centering
\caption{Fairness evaluation of pluralistic alignment across tasks, rewards, and aggregation strategies. $FI$ denotes the Fairness Index. Alignment scores are reported under multiple metrics; higher is better for all metrics except KL and Was. Both average (Avg AS) and minimum (Min AS) alignment scores are shown. For each column, the best values are highlighted in \textbf{bold} (lowest for KL and Was, highest otherwise).}

%\salma{Don't forget to add the inter- intra-group fairness index that we discussed}
\label{tab:fairness-alignment}
\renewcommand{\arraystretch}{1.1}
\setlength{\tabcolsep}{2pt}
\tiny
\begin{tabular}{l|| l | l | l || c c c c c c || c c c c c c || c c c c c c}
\toprule
\multirow{4}{*}{\textbf{Task}} &
\multirow{4}{*}{\makecell[l]{\textbf{Client}\\\textbf{Reward}}} &
\multirow{4}{*}{\textbf{Method}} &
\multirow{4}{*}{\textbf{Server Agg.}} &
\multicolumn{6}{c}{\textbf{Fairness Index ($FI$)}} &
\multicolumn{6}{c}{\textbf{Avg Alignment Score (Avg AS)}} &
\multicolumn{6}{c}{\textbf{Min Alignment Score (Min AS)}} \\
\cmidrule(l){5-10} \cmidrule(l){11-16} \cmidrule(l){17-22}
& & & &
\textbf{Was.} &
\textbf{Cos.} &
\textbf{KL} &
\textbf{Ken.} &
\textbf{Bor.} &
\textbf{Bin.} &
\textbf{Was.} &
\textbf{Cos.} &
\textbf{KL} &
\textbf{Ken.} &
\textbf{Bor.} &
\textbf{Bin.} &
\textbf{Was.} &
\textbf{Cos.} &
\textbf{KL} &
\textbf{Ken.} &
\textbf{Bor.} &
\textbf{Bin.} \\
\midrule

%==================== VALUE TASK ====================
\multirow{25}{*}{\rotatebox{90}{\textbf{Preference Prediction Task}}}
& \na & SFT & \na & 0.98 & 0.97 & 0.88 & 0.85 & 0.83 & 0.97 & 0.10 & 0.82 & 0.4 & 0.28 & 0.38 & 0.23 & 0.08 & 0.77 & 0.55 & 0.13 & 0.34 & 0.23 \\
\cmidrule(lr){2-22}
& \multirow{4}{*}{WassersteinReward}
    & \multirow{4}{*}{PPO} & Alpha  & \textbf{0.99} & \textbf{0.99} & \textbf{0.94} & \textbf{0.91} & \textbf{0.86} & \textbf{1.00} & 0.05 & 0.90 & 0.26 & 0.30 & 0.42 & 0.22 & \textbf{0.06} & \textbf{0.89} & \textbf{0.26} & 0.21 & 0.37 & 0.22 \\
&   &  & Min    & 0.98 & \textbf{0.99} & 0.93 & 0.87 & 0.79 & 0.90 & 0.05 & \textbf{0.91} & \textbf{0.22} & 0.42 & 0.44 & 0.27 & \textbf{0.06} & \textbf{0.89} & 0.27 & 0.34 & 0.41 & \textbf{0.28} \\
&   &  & Avg    & \textbf{0.99} & \textbf{0.99} & \textbf{0.94} & \textbf{0.91} & \textbf{0.86} & \textbf{1.00} & 0.05 & 0.90 & 0.26 & 0.30 & 0.42 & 0.22 & \textbf{0.06} & \textbf{0.89} & \textbf{0.26} & 0.21 & 0.37 & 0.22 \\
&   &  & Max    & \textbf{0.99} & \textbf{0.99} & 0.90 & 0.88 & 0.80 & 0.85 & \textbf{0.03} & \textbf{0.91} & 0.23 & \textbf{0.45} & \textbf{0.51} & \textbf{0.31} & 0.07 & \textbf{0.89} & 0.27 & \textbf{0.43} & \textbf{0.47} & 0.31 \\
\cmidrule(lr){2-22}
& \multirow{4}{*}{CosineReward}
    & \multirow{4}{*}{PPO} & Alpha  & \textbf{0.99} & \textbf{0.99} & 0.89 & \textbf{0.88} & \textbf{0.89} & \textbf{0.91} & \textbf{0.05} & 0.92 & 0.21 & 0.28 & 0.42 & 0.21 & \textbf{0.06} & 0.90 & 0.27 & 0.19 & 0.32 & 0.19 \\
&   &  & Min    & \textbf{0.99} & \textbf{0.99} & 0.90 & \textbf{0.88} & \textbf{0.89} & 0.80 & \textbf{0.05} & 0.92 & 0.22 & \textbf{0.34} & \textbf{0.45} & \textbf{0.28} & \textbf{0.06} & 0.90 & 0.28 & \textbf{0.21} & \textbf{0.34} & \textbf{0.22} \\
&   &  & Avg    & \textbf{0.99} & \textbf{0.99} & 0.89 & \textbf{0.88} & \textbf{0.89} & \textbf{0.91} & \textbf{0.05} & 0.92 & 0.21 & 0.28 & 0.42 & 0.21 & \textbf{0.06} & 0.90 & 0.27 & 0.19 & 0.32 & 0.19 \\
&   &  & Max    & \textbf{0.99} & \textbf{0.99} & \textbf{0.91} & 0.87 & 0.88 & 0.88 & \textbf{0.05} & \textbf{0.93} & \textbf{0.19} & 0.31 & 0.42 & 0.22 & \textbf{0.06} & \textbf{0.91} & \textbf{0.24} & 0.20 & \textbf{0.34} & 0.19 \\
\cmidrule(lr){2-22}
& \multirow{4}{*}{KLReward}
    & \multirow{4}{*}{PPO} & Alpha  & \textbf{0.99} & \textbf{0.99} & \textbf{0.92} & \textbf{0.92} & \textbf{0.90} & 0.78 & 0.06 & \textbf{0.92} & 0.19 & 0.40 & 0.50 & 0.29 & 0.07 & \textbf{0.90} & 0.24 & 0.18 & 0.38 & 0.22 \\
&   &  & Min    & \textbf{0.99} & \textbf{0.99} & 0.91 & 0.90 & 0.89 & \textbf{0.84} & 0.06 & 0.91 & \textbf{0.17} & \textbf{0.43} & \textbf{0.51} & \textbf{0.33} & 0.07 & 0.89 & \textbf{0.22} & \textbf{0.33} & \textbf{0.43} & 0.31 \\
&   &  & Avg    & \textbf{0.99} & \textbf{0.99} & 0.91 & 0.89 & \textbf{0.90} & 0.76 & 0.05 & 0.91 & 0.19 & 0.40 & 0.48 & 0.27 & 0.06 & 0.89 & 0.26 & 0.29 & 0.40 & \textbf{0.25} \\
&   &  & Max    & \textbf{0.99} & \textbf{0.99} & 0.91 & 0.90 & 0.86 & 0.75 & \textbf{0.04} & 0.91 & 0.19 & 0.33 & 0.40 & 0.20 & \textbf{0.05} & 0.88 & 0.24 & 0.22 & 0.33 & 0.19 \\
\cmidrule(lr){2-22}
& \multirow{4}{*}{KendallTauReward}
    & \multirow{4}{*}{PPO} & Alpha  & \textbf{0.99} & \textbf{0.99} & \textbf{0.96} & 0.90 & 0.71 & \textbf{0.91} & 0.07 & 0.75 & 0.48 & 0.43 & 0.38 & \textbf{0.29} & 0.08 & 0.72 & 0.55 & 0.34 & 0.36 & \textbf{0.28} \\
&   &  & Min    & \textbf{0.99} & \textbf{0.99} & 0.95 & 0.90 & 0.75 & \textbf{0.91} & 0.07 & 0.73 & 0.49 & \textbf{0.45} & \textbf{0.39} & \textbf{0.29} & 0.08 & 0.71 & 0.54 & 0.37 & 0.36 & \textbf{0.28} \\
&   &  & Avg    & \textbf{0.99} & \textbf{0.99} & 0.94 & 0.90 & \textbf{0.76} & \textbf{0.91} & 0.07 & 0.74 & 0.48 & \textbf{0.45} & \textbf{0.39} & \textbf{0.29} & 0.08 & 0.71 & 0.54 & \textbf{0.38} & \textbf{0.37} & \textbf{0.28} \\
&   &  & Max    & \textbf{0.99} & \textbf{0.99} & 0.94 & \textbf{0.92} & 0.73 & \textbf{0.91} & \textbf{0.06} & \textbf{0.76} & \textbf{0.44} & 0.44 & 0.38 & 0.28 & \textbf{0.07} & \textbf{0.73} & \textbf{0.50} & 0.37 & 0.35 & \textbf{0.28} \\
\cmidrule(lr){2-22}
& \multirow{4}{*}{BordaReward}
    & \multirow{4}{*}{PPO} & Alpha  & \textbf{0.99} & \textbf{0.99} & 0.96 & \textbf{0.89} & \textbf{0.71} & 0.91 & \textbf{0.08} & 0.73 & 0.50 & 0.43 & \textbf{0.39} & \textbf{0.29} & \textbf{0.09} & 0.69 & 0.57 & 0.35 & \textbf{0.36} & \textbf{0.28} \\
&   &  & Min    & \textbf{0.99} & \textbf{0.99} & \textbf{0.98} & 0.86 & 0.69 & \textbf{0.92} & 0.09 & 0.73 & 0.52 & \textbf{0.44} & \textbf{0.39} & 0.28 & 0.10 & \textbf{0.70} & 0.58 & \textbf{0.37} & \textbf{0.36} & \textbf{0.28} \\
&   &  & Avg    & \textbf{0.99} & \textbf{0.99} & 0.97 & \textbf{0.89} & \textbf{0.71} & 0.91 & 0.09 & 0.73 & 0.51 & 0.42 & 0.38 & \textbf{0.29} & 0.10 & \textbf{0.70} & 0.58 & 0.35 & \textbf{0.36} & \textbf{0.28} \\
&   &  & Max    & \textbf{0.99} & \textbf{0.99} & 0.97 & \textbf{0.89} & \textbf{0.71} & 0.90 & \textbf{0.08} & \textbf{0.74} & \textbf{0.49} & 0.44 & \textbf{0.39} & \textbf{0.29} & \textbf{0.09} & \textbf{0.70} & \textbf{0.56} & 0.36 & \textbf{0.36} & \textbf{0.28} \\
\cmidrule(lr){2-22}
& \multirow{4}{*}{BinaryReward}
    & \multirow{4}{*}{PPO} & Alpha  & \textbf{0.99} & \textbf{0.99} & 0.96 & \textbf{0.89} & \textbf{0.68} & \textbf{1.00} & \textbf{0.08} & \textbf{0.74} & \textbf{0.52} & \textbf{0.42} & \textbf{0.38} & 0.28 & \textbf{0.10} & \textbf{0.70} & 0.60 & 0.34 & 0.35 & \textbf{0.28} \\
&   &  & Min    & \textbf{0.99} & \textbf{0.99} & \textbf{0.98} & 0.86 & 0.66 & \textbf{1.00} & 0.10 & 0.70 & 0.70 & 0.37 & 0.37 & 0.28 & 0.11 & 0.66 & 0.77 & 0.30 & 0.35 & \textbf{0.28} \\
&   &  & Avg    & \textbf{0.99} & \textbf{0.99} & 0.97 & \textbf{0.89} & 0.67 & \textbf{1.00} & 0.09 & 0.73 & 0.54 & 0.42 & \textbf{0.38} & \textbf{0.29} & \textbf{0.10} & \textbf{0.70} & \textbf{0.59} & \textbf{0.35} & \textbf{0.36} & \textbf{0.28} \\
&   &  & Max    & \textbf{0.99} & \textbf{0.99} & 0.97 & \textbf{0.89} & \textbf{0.68} & \textbf{1.00} & \textbf{0.08} & 0.73 & 0.56 & 0.41 & \textbf{0.38} & 0.28 & \textbf{0.10} & \textbf{0.70} & 0.64 & 0.33 & 0.34 & \textbf{0.28} \\

\midrule

%==================== ORDER TASK ====================
\multirow{13}{*}{\rotatebox{90}{\textbf{Preference Ranking Task}}}
& \na & SFT & \na & \na & \na & \na & 0.89 & 0.87 & 0.83 & \na & \na & \na & 0.38 & 0.50 & 0.31 & \na & \na & \na & 0.25 & 0.41 & 0.27 \\
\cmidrule(lr){2-22}
& \multirow{4}{*}{KendallTauReward}
    & \multirow{4}{*}{PPO} & Alpha  & \na    & \na    & \na    & \textbf{0.92} & 0.81 & 0.97 & \na    & \na    & \na    & \textbf{0.58} & 0.47 & \textbf{0.36} & \na    & \na    & \na    & \textbf{0.47} & 0.42 & \textbf{0.31} \\
&   &  & Min    & \na    & \na    & \na    & \textbf{0.92} & 0.81 & \textbf{0.99} & \na    & \na    & \na    & 0.52 & 0.46 & 0.30 & \na    & \na    & \na    & 0.43 & 0.41 & 0.28 \\
&   &  & Avg    & \na    & \na    & \na    & \textbf{0.92} & 0.82 & 0.90 & \na    & \na    & \na    & 0.50 & 0.48 & 0.33 & \na    & \na    & \na    & 0.40 & 0.40 & 0.28 \\
&   &  & Max    & \na    & \na    & \na    & 0.91 & \textbf{0.88} & 0.80 & \na    & \na    & \na    & 0.47 & \textbf{0.53} & 0.35 & \na    & \na    & \na    & 0.35 & \textbf{0.44} & 0.28 \\
\cmidrule(lr){2-22}
& \multirow{4}{*}{BordaReward}
    & \multirow{4}{*}{PPO} & Alpha  & \na    & \na    & \na    & \textbf{0.94} & \textbf{0.95} & 0.86 & \na    & \na    & \na    & \textbf{0.53} & \textbf{0.61} & \textbf{0.39} & \na    & \na    & \na    & 0.34 & 0.45 & 0.28 \\
&   &  & Min    & \na    & \na    & \na    & 0.92 & 0.91 & \textbf{0.89} & \na    & \na    & \na    & 0.47 & 0.53 & 0.30 & \na    & \na    & \na    & \textbf{0.36} & 0.44 & 0.28 \\
&   &  & Avg    & \na    & \na    & \na    & 0.93 & 0.92 & 0.86 & \na    & \na    & \na    & 0.49 & 0.58 & \textbf{0.39} & \na    & \na    & \na    & 0.35 & \textbf{0.47} & \textbf{0.31} \\
&   &  & Max    & \na    & \na    & \na    & 0.91 & 0.92 & 0.78 & \na    & \na    & \na    & 0.45 & 0.54 & 0.32 & \na    & \na    & \na    & 0.34 & 0.45 & 0.28 \\
\cmidrule(lr){2-22}
& \multirow{4}{*}{BinaryReward}
    & \multirow{4}{*}{PPO} & Alpha  & \na    & \na    & \na    & \textbf{0.91} & 0.89 & 0.79 & \na    & \na    & \na    & \textbf{0.49} & 0.53 & \textbf{0.35} & \na    & \na    & \na    & 0.33 & 0.42 & 0.25 \\
&   &  & Min    & \na    & \na    & \na    & 0.90 & 0.83 & \textbf{0.90} & \na    & \na    & \na    & \textbf{0.49} & 0.49 & 0.34 & \na    & \na    & \na    & \textbf{0.39} & 0.41 & \textbf{0.28} \\
&   &  & Avg    & \na    & \na    & \na    & \textbf{0.91} & 0.90 & 0.79 & \na    & \na    & \na    & \textbf{0.49} & 0.53 & \textbf{0.35} & \na    & \na    & \na    & 0.37 & \textbf{0.44} & \textbf{0.28} \\
&   &  & Max    & \na    & \na    & \na    & \textbf{0.91} & \textbf{0.91} & 0.79 & \na    & \na    & \na    & 0.47 & \textbf{0.54} & \textbf{0.35} & \na    & \na    & \na    & 0.35 & \textbf{0.44} & \textbf{0.28} \\

\bottomrule
\end{tabular}
\end{table*}

\subsection{Preference Probability Prediction Task Results and Analysis}

The SFT baseline demonstrates suboptimal performance with fairness indices
ranging from 0.83--0.98 and consistently lower alignment scores, highlighting
the need for preference-based alignment. Detailed quantitative results are
shown in \autoref{tab:fairness-alignment} across multiple reward functions in
the Value task. 

\textbf{Reward Function Analysis:}
Distance-based rewards (Wasserstein, Cosine, KL) substantially outperform
ranking-based approaches (Kendall, Borda, Binary). Wasserstein and Cosine
rewards with \textsc{Alpha} aggregation achieve near-optimal fairness
($FI \approx 0.99$) while maintaining strong average alignment scores
($\mathrm{Avg\,AS} \approx 0.90$--0.95). The minimum alignment scores, crucial
for ensuring that no group is left behind, remain competitive
($\mathrm{Min\,AS} \approx 0.89$--0.94), demonstrating an effective
fairness–performance balance.

\textbf{Dynamic Alpha Aggregation Strategy Impact:}
Across all distance-based rewards, our adaptive \textsc{Alpha} aggregation
consistently achieves superior fairness indices (often $FI = 0.99$) while
preserving competitive average and minimum alignment scores. Compared to
static strategies (\textsc{Min}, \textsc{Avg}, \textsc{Max}), \textsc{Alpha}
shows two key benefits: (i) it avoids the fairness degradation and
worst-group collapse observed with \textsc{Max}, which can push up average
alignment at the cost of marginalized groups; and (ii) it improves
worst-group performance relative to \textsc{Avg}, leading to higher
$\mathrm{Min\,AS}$ at comparable or better $FI$. This behavior stems from
dynamically reweighting groups based on historical alignment, allowing the
server to upweight under-served groups without sacrificing overall utility.

\textbf{Recommendations:}
For probability-based preference prediction tasks, we recommend using
Wasserstein or Cosine rewards combined with \textsc{Alpha} aggregation as the
default configuration. This combination consistently achieves near-perfect
fairness ($FI \approx 0.99$) while maintaining strong average and minimum
alignment scores across groups. In settings where heightened sensitivity to
distributional mismatch is desired, KL-based rewards with \textsc{Alpha}
aggregation remain competitive but may induce slightly larger fairness
variance. Overall, Wasserstein/Cosine + \textsc{Alpha} provides the most
favorable fairness–performance trade-off for modeling calibrated group-level
preference probabilities.

\begin{figure*}[t]
  \centering
  \includegraphics[width=\textwidth]{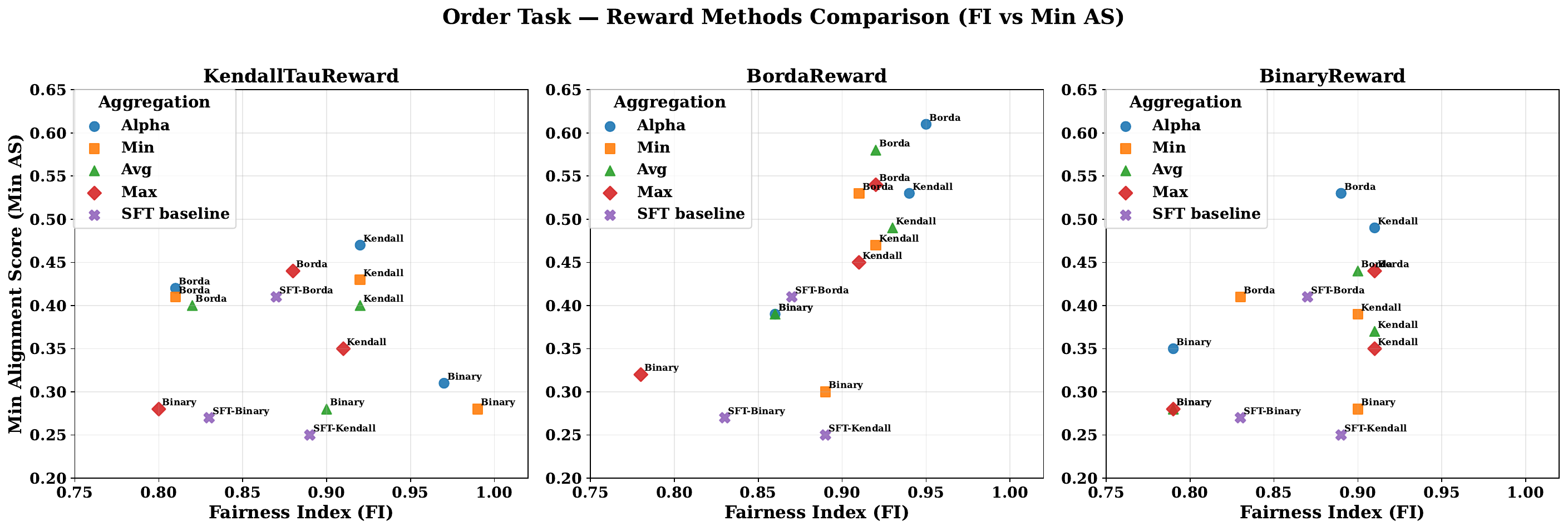}
  \caption{\textbf{Order task — FI vs.~Min AS (worst–group performance) with reward-as-aggregation.}
  Each subplot fixes \emph{one reward as the main aggregation metric} on the server—
  \textit{KendallTauReward} (left), \textit{BordaReward} (middle), and \textit{BinaryReward} (right)—
  and then measures its effect on the three ranking metrics (Kendall, Borda, Binary).
  Points are server aggregation strategies (\textsc{Alpha}, \textsc{Min}, \textsc{Avg}, \textsc{Max});
  SFT baselines are purple crosses. We use the \textbf{minimum alignment score (Min AS)} on the y-axis
  because it reflects the performance of the \emph{worst-served group}, while the x-axis shows the
  Fairness Index (FI).}
  \label{fig:order-task-comparison}
\end{figure*}

\subsection{Preference Ranking Task Results and Analysis}

The Order task evaluation focuses on ranking-based metrics (Kendall Tau, Borda, Binary). More details are shown in \autoref{tab:fairness-alignment}. The SFT baseline, trained on population-averaged preferences, shows particularly poor performance in ranking tasks with fairness indices of 0.83-0.89 and substantially lower alignment scores (0.31-0.50 average, 0.25-0.41 minimum). This performance degradation in ranking tasks underscores how averaged preference training fails to capture the nuanced ordering preferences that vary significantly across demographic groups.

\textbf{Ranking Reward Analysis:} Alpha aggregation with Kendall Tau rewards achieves the highest fairness index (0.92) and superior average alignment scores (0.58) compared to the SFT baseline (0.38). Borda rewards demonstrate the strongest overall performance, reaching fairness indices up to 0.95 with alpha aggregation and achieving the highest average alignment scores (0.61).

Figure~\ref{fig:order-task-comparison} demonstrates that \textbf{\textsc{Alpha} aggregation (blue circles) consistently provides the best fairness–performance trade-off,}
  occupying or approaching the upper-right quadrant ($\mathrm{FI}>0.9$, $\mathrm{Min\,AS}>0.3$) in all panels.
  For \textit{KendallTauReward}, \textsc{Alpha} attains high FI and strong worst-group performance across metrics
  (Kendall: $\mathrm{FI}\!\approx\!0.92$, $\mathrm{Min\,AS}\!\approx\!0.47$; Borda: $0.81,0.42$; Binary: $0.97,0.31$),
  whereas \textsc{Min} pushes FI high on Binary ($\approx\!0.99$) but \emph{hurts} Min AS ($\approx\!0.28$).
  For \textit{BordaReward}, \textsc{Alpha} achieves the \emph{highest} Min AS across metrics (Kendall $\approx\!0.53$,
  Borda $\approx\!0.61$, Binary $\approx\!0.39$) with top/tied FI (Kendall $\approx\!0.94$, Borda $\approx\!0.95$,
  Binary $\approx\!0.86$).
  For \textit{BinaryReward}, \textsc{Alpha} again yields the largest Min AS (Kendall $\approx\!0.49$, Borda
  $\approx\!0.53$, Binary $\approx\!0.35$) with competitive FI, outperforming \textsc{Min}/\textsc{Avg}/\textsc{Max}
  in protecting the worst-served group.
  Across panels, the SFT baseline underperforms, reinforcing the need for federated preference alignment.
  Overall, the visualization shows that \textsc{Alpha} most effectively resolves the fairness–performance tension
  by \emph{maximizing worst-group (Min AS) performance at high FI}.
% visualizes the fairness-performance trade-off for preference ranking tasks using Kendall Tau rewards. Our adaptive alpha aggregation (blue circles) consistently occupies the optimal high-fairness, high-performance quadrant (FI $>$ 0.9, Min AS $>$ 0.3), with Binary rewards achieving near-perfect fairness (FI $\approx$ 0.97) and the highest minimum alignment score (Min AS $\approx$ 0.31). Other aggregation strategies show suboptimal positioning: Min aggregation clusters at lower fairness indices, Max aggregation demonstrates inconsistent performance, and Average aggregation achieves mediocre results in both dimensions. The SFT baseline (purple crosses) underperforms across all metrics, confirming the necessity of federated preference alignment. This visualization validates that adaptive alpha aggregation effectively resolves the fairness-performance dilemma through dynamic group prioritization.

\textbf{Dynamic Alpha Aggregation Strategy Impact:} Across all ranking rewards, alpha aggregation maintains competitive performance while consistently achieving better fairness indices than alternative aggregation strategies. Importantly, our evaluation of minimum alignment scores reveals that alpha aggregation successfully prevents the marginalization of lowest-performing groups, maintaining minimum scores (0.31-0.47) that are competitive with or superior to other approaches, while simultaneously achieving higher average performance.

\textbf{Recommendations:} For order-based tasks, we recommend Borda rewards with alpha aggregation, which provides the optimal balance between fairness (0.94-0.95) and alignment performance (0.53-0.61 average). This combination effectively captures group ranking preferences while maintaining equitable treatment across demographic groups.

\subsection{Overall Assessment}

Our adaptive alpha aggregation demonstrates superior performance across both task types, consistently achieving the highest fairness indices while maintaining competitive alignment scores. The approach successfully addresses the critical challenge of preventing any group from being left behind, as evidenced by competitive minimum alignment scores across all evaluation scenarios. These results validate our hypothesis that adaptive weighting based on historical alignment performance provides an effective mechanism for achieving equitable federated learning in preference alignment tasks.
\section{Limitations and Future Work}

While our study demonstrates the effectiveness of adaptive alpha aggregation for pluralistic alignment, several areas present natural directions for further research.

\textbf{Underlying RL Framework.} Our current implementation relies on PPO. While effective, PPO can be computationally expensive. Future work should explore more resource-efficient alternatives such as GRPO or DPO, which would allow testing the aggregation strategy across different optimization paradigms and at larger scales.

\textbf{Model and Dataset Scope.} We evaluate on Gemma-2B-it using the Pew Research Global Attitudes dataset, which may be relatively conducive to cross-group alignment. Broader validation on base models and domains with more adversarial or conflicting preferences would provide a stronger stress test of the method’s robustness.

\textbf{Task Generalization.} Our experiments focus on multiple-choice Q\&A tasks, which offer a controlled setting for evaluation. Extending the framework to diverse tasks such as summarization, dialogue, or code generation would demonstrate its wider applicability and highlight how aggregation impacts more open-ended alignment scenarios.

These considerations do not detract from our main contribution—a systematic evaluation framework with a novel adaptive aggregation scheme—but rather open exciting avenues for expanding its applicability across models, datasets, and tasks.

\section{Conclusion}\label{sec:conclusion}

% We evaluated adaptive alpha aggregation scheme successfully achieves equitable LLM alignment in federated settings by dynamically adjusting preference weights based on historical alignment performance. Our evaluations confirm this method consistently improves fairness over standard techniques without compromising alignment quality, crucially preventing the marginalization of underrepresented groups. The approach thus provides a robust and practical solution for developing more pluralistic and fairly aligned LLMs in decentralized environments.
% In conclusion, our work addresses the critical challenge of evaluating LLM alignment in decentralized, federated environments. We have demonstrated that the choice of aggregation technique is not a trivial detail but a fundamental evaluation protocol that directly shapes a model's fairness and performance. Our systematic evaluation of standard aggregation methods, alongside the introduction of an adaptive scheme, provides a clear framework for assessing the trade-offs between alignment and fairness. The results show that our proposed Adaptive Alpha Aggregation achieves a superior balance, offering a practical path toward developing truly pluralistic and equitably aligned LLMs. This research contributes a valuable evaluation methodology to the field and opens up new avenues for future work, including applying this framework to a broader range of tasks and model architectures.

% \section{Conclusion}\label{sec:conclusion}

This work tackles the challenge of aligning LLMs with diverse human preferences in decentralized, federated settings. We showed that \emph{how} group feedback is aggregated is not a minor implementation detail, but a central part of the evaluation protocol that directly governs both fairness and overall alignment performance. Building on PluralLLM's group-specific preference modeling, we proposed an evaluation framework that spans probability prediction and ranking tasks, multiple reward functions, and several aggregation baselines.

Within this framework, our Adaptive Alpha Aggregation dynamically reweights groups based on their historical alignment performance, consistently improving cross-group fairness while maintaining competitive alignment scores. In particular, it raises the performance of the worst-served groups without sacrificing overall utility, providing a practical path toward more pluralistic and equitably aligned LLMs in federated RLHF. This research contributes a valuable evaluation methodology to the field and opens up new avenues for future work, including applying this framework to a broader range of tasks and model architectures.

\section*{Acknowledgments}
This work is supported by the U.S. National Science Foundation (NSF) under grant number 2339266.

% \begin{ack}
% This work is supported by the U.S. National Science Foundation (NSF) under grant number 2339266.
% \end{ack}

% \newpage

% \bibliographystyle{ACM-Reference-Format}
% \bibliographystyle{IEEEtran}
% \bibliographystyle{iclr2026_conference}
\bibliographystyle{plainnat}   % or unsrtnat / abbrvnat
\bibliography{reference}

% \appendix
% \section{Appendix}
\newpage
\appendix
% \newpage
\appendix
% \newpage
% \renewcommand{\thesection}{\arabic{section}}
% \section{Appendices}

\section{Experiment Configurations and Hyperparameters}\label{appendix:setup}

Our experimental setup begins with supervised fine-tuning (SFT) as outlined in Table~\ref{tab:sft-hparams}. We use the Gemma-2-2b-it model as our base, employing LoRA adaptation with rank 16 for efficient parameter updates. The SFT training utilizes a cosine learning rate scheduler with warmup and is conducted for a single epoch to establish our baseline model.
% =========================
% SFT Hyperparameters
% =========================
\begin{table}[h]
\centering
\caption{SFT configuration and hyperparameters.}
\label{tab:sft-hparams}
\begin{tabular}{l l}
\toprule
\textbf{Hyperparameter} & \textbf{Value} \\
\midrule
\multicolumn{2}{l}{\textit{Model}} \\
Base model & \texttt{google/gemma-2-2b-it} \\
Precision & BF16 \\
\addlinespace[2pt]
\multicolumn{2}{l}{\textit{Data / Task}} \\
Train/valid split & 80/20\% \\
Max sequence length & 500 (include prompt and response) \\
\addlinespace[2pt]
\multicolumn{2}{l}{\textit{LoRA Adapter}} \\
Rank (\(r\)) & 16 \\
Alpha & 32 \\
Dropout & 0.05 \\
\addlinespace[2pt]
\multicolumn{2}{l}{\textit{Optimization}} \\
Batch size (per device) & 16 \\
Gradient accumulation steps & 4 \\
Learning rate & \(5\times10^{-5}\) \\
Scheduler & cosine \\
Warmup steps & 150 \\
Weight decay & 0.01 \\
\addlinespace[2pt]
\multicolumn{2}{l}{\textit{Training}} \\
Epochs & 1 \\
\bottomrule
\end{tabular}
\end{table}

% =========================
% PPO Hyperparameters
% =========================
\begin{table}[h]
\centering
\caption{PPO configuration and hyperparameters (policy and value models initialized from the SFT model).}
\label{tab:ppo-hparams}
\begin{tabular}{l l}
\toprule
\textbf{Hyperparameter} & \textbf{Value} \\
\midrule
\multicolumn{2}{l}{\textit{General}} \\
Policy model & Gemma 2 SFT model \\
Value model & Gemma 2 SFT model \\
\addlinespace[2pt]
\multicolumn{2}{l}{\textit{Model / Quantization}} \\
Quantization & 4-bit (\texttt{nf4}, double-quant = True) \\
Compute dtype & BF16 \\
Attention implementation & eager \\
\addlinespace[2pt]
\multicolumn{2}{l}{\textit{LoRA (PEFT)}} \\
Rank (\(r\)) & 32 \\
Alpha & 32 \\
Dropout & 0.05 \\
\addlinespace[2pt]
\multicolumn{2}{l}{\textit{Optimization}} \\
Per-device train batch size & 4 \\
Gradient accumulation steps & 24 \\
Learning rate & \(1\times10^{-5}\) \\
Optimizer & AdamW \\
Weight decay & 0.0 \\
Scheduler & linear \\
\addlinespace[2pt]
\multicolumn{2}{l}{\textit{PPO Trainer}} \\
PPO epochs & 2 \\
Mini-batches & 8 \\
Per-device eval batch size & 32 \\
Response length & 42 \\
Temperature & 0.6 \\
KL coefficient & 0.05 \\
Clip range & 0.2 \\
Clip range (value) & 0.2 \\
Value loss coefficient (\(v_f\)) & 0.2  \\
Discount factor (\(\gamma\)) & 1.0 \\
GAE lambda (\(\lambda\)) & 0.95 \\
Reward whitening & Per rollout (before PPO update)\footnotemark \\
\addlinespace[2pt]
\bottomrule
\end{tabular}
\end{table}
\footnotetext{Rewards are whitened over each rollout before PPO updates.}

As summarized in Table~\ref{tab:ppo-hparams}, both the policy and value models in PPO are initialized from the SFT model. During training, we employ two distinct prompt formats for evaluation: a preference probability prediction task requiring models to assign probability scores to all options, and a preference ranking task requiring complete ordinal ranking from most to least preferred (see Figures~\ref{fig:prob-prompt} and~\ref{fig:rank-prompt}). Our implementation builds upon the Hugging Face TRL library \cite{vonwerra2022trl}. All experiments were conducted on 3 nodes, each equipped with A100 GPUs, Intel(R) Xeon(R) Gold 6326 CPUs @ 2.90GHz, and 256GB RAM.

% \input{appendix/prompt}

% \input{appendix/metric}

% \input{appendix/proof}

%\newpage 
% \section{Comparison between Value task and Order task }\label{app:allresults}

% Detailed quantitative results are shown  in \autoref{tab:fairness-alignment} across multiple reward functions.

%\newpage

% \pagebreak

\end{document}